\documentclass{article}

\usepackage[preprint]{neurips_2025}

\usepackage[utf8]{inputenc} 
\usepackage[T1]{fontenc}    
\usepackage[dvipsnames,table,xcdraw]{xcolor}
\usepackage[pagebackref,breaklinks,colorlinks,linkcolor=BrickRed,citecolor=RoyalBlue]{hyperref}
\usepackage{url}            
\usepackage{booktabs}       
\usepackage{amsfonts}       
\usepackage{nicefrac}       
\usepackage{microtype}      
\usepackage{xcolor}         

\usepackage{colortbl}

\usepackage{amsmath}
\usepackage{amssymb}
\usepackage{pifont}
\newcommand{\cmark}{\ding{51}}%

\usepackage{textcomp}

\usepackage{booktabs}
\usepackage{multirow}
\usepackage{makecell}

\usepackage{algorithm}
\usepackage{algorithmic}

\usepackage{overpic} 
\usepackage{contour}
\usepackage{color}

\usepackage{float} 
\usepackage{subcaption}

\newcommand{\nameofmethod}{MedSeg-R}
\title{\nameofmethod: Dual‑Stage Unified Overlap‑Aware Segmentation for Medical Images}
\title{\nameofmethod: Medical Image Segmentation with \\ Clinical Reasoning}

\author{
    Hao Shao\textsuperscript{\thanks{Email at shaoh@mail.nankai.edu.cn}}, 
    Qibin Hou\textsuperscript{\thanks{Corresponding authors, email at: houqb@nankai.edu.cn}}\\
    VCIP, School of Computer Science, Nankai University \\
      \texttt{Project page: \url{https://github.com/haoshao-nku/MedSeg-R.git}} \\
    }

\begin{document}

\maketitle

\begin{abstract}
Medical image segmentation is challenging due to overlapping anatomies with ambiguous boundaries and a severe imbalance between the foreground and background classes, which particularly affects the delineation of small lesions. 
Existing methods, including encoder–decoder networks and prompt-driven variants of the Segment Anything Model (SAM), rely heavily on local cues or user prompts and lack integrated semantic priors, thus failing to generalize well to low-contrast or overlapping targets.
To address these issues, we propose~\nameofmethod{}, a lightweight, dual-stage framework inspired by inspired by clinical reasoning.
Its cognitive stage interprets medical report into structured semantic priors (location, texture, shape), which are fused via transformer block. 
In the perceptual stage, these priors modulate the SAM backbone: spatial attention highlights likely lesion regions, dynamic convolution adapts feature filters to expected textures, and deformable sampling refines spatial support.
By embedding this fine-grained guidance early, \nameofmethod{} disentangles inter-class confusion and amplifies minority-class cues, greatly improving sensitivity to small lesions.
In challenging benchmarks, \nameofmethod{} produces large Dice improvements in overlapping and ambiguous structures, demonstrating  plug-and-play compatibility with SAM-based systems.
\end{abstract}

\section{Introduction}

Medical image segmentation is a cornerstone of numerous clinical workflows, enabling the precise delineation of organs, lesions, and anatomical structures across imaging modalities such as CT, MRI, and X-ray~\cite{ma2022fast,shao2024polyper,gaggion2022improving}.
Its pixel-level accuracy underpins a range of critical downstream tasks, including quantitative analysis, radiotherapy planning, and longitudinal disease monitoring~\cite{schipaanboord2019evaluation,iqbal2023ldmres}.
Despite substantial progress, two persistent challenges continue to hinder robust and generalizable segmentation:
(1) inter-class overlap, where spatially adjacent anatomical regions exhibit highly similar intensity and texture patterns, resulting in ambiguous boundaries; and
(2) small object segmentation, where extreme class imbalance leads to underrepresentation and unreliable detection of subtle but clinically significant findings.

Early convolutional models, including U-Net~\cite{ronneberger2015u} and its attention-based~\cite{oktay2018attention}, improve multi-scale feature integration but are still limited by local receptive fields.
Subsequent Transformer-based methods~\cite{chen2021transunet,zheng2021rethinking} introduce global context modeling, improving anatomical reasoning.
However, both families of methods still struggle with overlapping structures and small objects.

\begin{figure}[t]
    \centering
      \begin{overpic}[width=1\textwidth]{{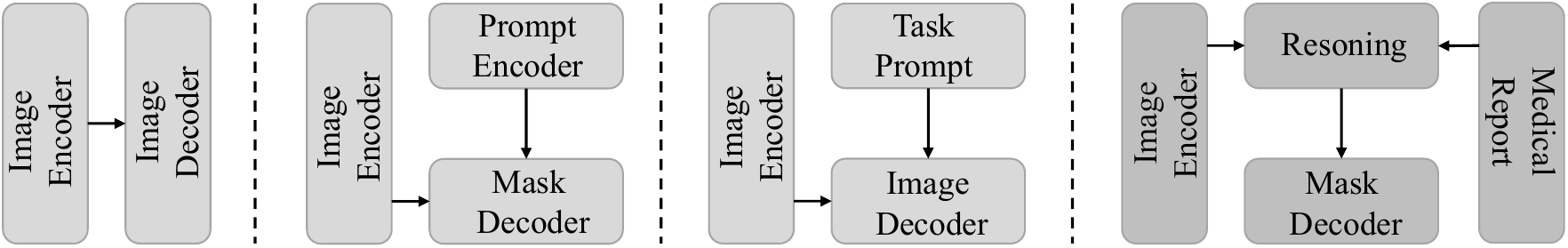}}
    \put(5.5, 17){{\textcolor{black}{(a)}}}
    \put(29.5, 17){{\textcolor{black}{(b)}}}
    \put(54.5, 17){{\textcolor{black}{(c)}}}
    \put(84, 17){{\textcolor{black}{(d)}}}
  \end{overpic}
    \caption{: Comparison of~\nameofmethod{} with existing approaches. (a) CNN/Transformer-based models, (b) prompt-driven SAM variants, (c) cross-dataset learning methods, and (d) our~\nameofmethod{}.
    Different from others,~\nameofmethod{} adopts a unique two-stage paradigm inspired by clinical reasoning, where structured priors (e.g., location, texture, shape) explicitly guide the visual feature extraction process.
    }    
    \label{fig:intro}
    \vspace{-0.5cm}
\end{figure}
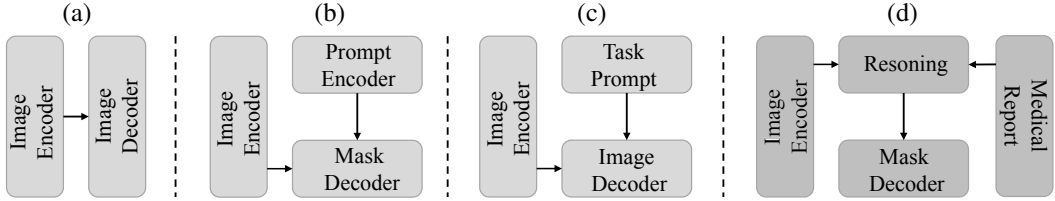

To pursue more generalizable medical image segmentation, two prominent paradigms have emerged: \textbf{prompt-driven modeling} and \textbf{cross-dataset learning}.
The first paradigm, \textbf{prompt-driven modeling}, is exemplified by the Segment Anything Model (SAM), which introduces a flexible prompt-based segmentation mechanism with strong generalization capabilities. 
Its medical adaptations, such as MedSAM~\cite{ma2024segment} and SAM-Med2D~\cite{cheng2023sam}, leverage domain-specific fine-tuning to deliver promising results across diverse medical imaging tasks.
However, these models still rely heavily on explicit prompts and lack semantic reasoning capabilities, limiting their performance in scenarios involving ambiguous boundaries or subtle lesions.
The second paradigm, \textbf{cross-dataset learning}, is represented by frameworks such as UniSeg~\cite{ye2023uniseg} and UniverSeg~\cite{butoi2023universeg}, which aim to generalize across modalities and anatomical targets via large-scale, cross-domain training.
Although effective in reducing overfitting to specific datasets, these models remain susceptible to domain shifts, require substantial manual annotation, and struggle with semantic ambiguity and class imbalance—especially in the case of small or overlapping structures.

All prevailing segmentation paradigms, including convolutional, Transformer-based, prompt driven, and cross‐dataset methods, share a fundamental flaw: They are difficult to fully exploit structured clinical priors, such as organ location, characteristic texture patterns, and lesion shapes, throughout the processing pipeline. They are mainly based on raw image features as shown in Fig.~\ref{fig:intro}.
In contrast, clinicians interpret medical images using extensive prior knowledge.
Even before identifying a structure, they recall expected anatomical locations, typical tissue appearances, and the morphological signatures of common pathologies.
These structured priors guide their attention to diagnostically relevant regions, enabling them to resolve overlapping boundaries and detect subtle abnormalities that purely image-driven approaches often overlook.

SEG-SAM~\cite{huang2024seg} takes a meaningful step toward leveraging clinical priors by introducing structured textual cues such as object location, object texture patterns, and object shapes through a dual-decoder design.
However, it encodes this information as a single global sentence embedding, offering only coarse semantic guidance.
The added decoder also increases the complexity of the system without directly improving the feature representation process.
As a result, SEG-SAM underutilizes fine-grained priors. 


Motivated by the need to better utilize structured clinical priors, we propose~\nameofmethod{}, a two-stage framework tailored for medical image segmentation as shown in Fig~\ref{fig:baseline}.
Unlike SEG-SAM~\cite{huang2024seg}, which encodes prior knowledge into a single global sentence vector, our method focuses on how to fully utilize fine-grained priors, such as object location, object texture patterns, and object shapes, while avoiding additional architectural complexity.
Specifically, \nameofmethod{} adopts a two-stage design inspired by clinical reasoning.
In the cognitive stage, it parses radiology-style descriptions into structured semantic attributes and fuses them via cross-attention into a unified prior. 
In the perceptual stage, this prior adaptively modulates the visual backbone of SAM, directing attention to the likely anatomical regions, enhancing the sensitivity to expected textures and aligning the receptive fields with the morphology of the lesion.
By embedding fine-grained guidance directly into the feature extraction process,~\nameofmethod{} avoids redundant decoding paths while improving segmentation accuracy for small, ambiguous, and overlapping structures.
Our contributions are three-fold.
\begin{itemize}
\item We propose a novel two-stage segmentation paradigm inspired by clinical reasoning, consisting of a cognitive stage that parses diagnostic descriptions into structured semantic attributes and a perceptual stage that uses these attributes to guide feature extraction and resolve ambiguities in complex anatomical boundaries.
\item We design lightweight and interpretable modules that inject structured clinical priors into the SAM backbone via attention and dynamic convolution mechanisms, enhancing segmentation accuracy for small and overlapping structures.
\item  We demonstrate that~\nameofmethod{} significantly outperforms state-of-the-art SAM-based methods across overlapping and small-object scenarios.
On the SAM-Med2D benchmark, it achieves consistent gains under both point and box prompts, improving small-object Dice scores by up to \textbf{+1.99\%} and \textbf{+1.76\%}, respectively. 
Furthermore,~\nameofmethod{} maintains superior performance even without any prompts, confirming its strong generalization and reduced reliance on external guidance.
On the Synapse dataset, it outperforms SEG-SAM by \textbf{+2.62\%} in overall Dice, with substantial improvements on challenging organs like the gallbladder (\textbf{+5.03\%}) and pancreas (\textbf{+2.22\%}).

\end{itemize}

\section{Related Work}

\textbf{Traditional medical image segmentation.}
Medical image segmentation requires balancing fine-grained boundary delineation with global semantic consistency, particularly in the presence of anatomical overlap and small lesions.
U-Net~\cite{ronneberger2015u} pioneers encoder-decoder architectures with skip connections, forming the basis for numerous subsequent methods.
Enhancements such as Attention U-Net~\cite{oktay2018attention} and UNet++~\cite{zhou2018unet++} introduce attention mechanisms and nested connectivity to improve feature integration and localization.
Volumetric variants like 3D U-Net~\cite{cciccek20163d} and nnU-Net~\cite{isensee2018nnu} extend segmentation to 3D modalities, enabling context-aware predictions across slices.
However, the above models share inherent limitations: Their fixed receptive fields constrain long-range dependency modeling, and their performance deteriorates in regions with low contrast or overlapping anatomical structures.
%
%
Transformer-based architectures~\cite{vaswani2017attention} mitigate these limitations by introducing global self-attention.
TransUNet~\cite{chen2021transunet} combines CNN encoders with Transformer modules to fuse local and global information, while SETR~\cite{zheng2021rethinking} explored pure transformer decoders.
Subsequent models like CoTr~\cite{xie2021cotr}, TransBTS~\cite{wenxuan2021transbts}, and Swin-UNet~\cite{cao2022swin} further advance spatial coherence and multi-scale representation.
However, these methods still have difficulties in handling fine-grained object boundaries and small object sensitivity due to insufficient supervision and over-smoothing of representations.

\textbf{Universal medical segmentation.}
To promote generalization across organs and modalities, one of the paradigms of universal segmentation frameworks is to learn shared representations through cross-task training.
UniSeg~\cite{ye2023uniseg} and UniverSeg~\cite{butoi2023universeg} achieve cross-dataset generalization by aligning anatomy-aware features through multi-domain supervision, enabling zero-shot segmentation of unseen structures.
However, their reliance on large-scale labeled corpora and their vulnerability to domain shifts limit real-world applicability.
%
%
To improve task-awareness, Iris~\cite{gao2025show} introduces a reference-guided framework that extracts relevant semantics from support examples.
However, its dependence on high-quality references reduces robustness in low-resource settings or under noisy annotations.
In volumetric contexts, 3DU$^2$-Net~\cite{huang20193d} and S2VNet~\cite{ding2024clustering} introduce architectural strategies to reduce redundancy and memory costs through separable convolutions or centroid-guided propagation.
These methods struggle to maintain accurate lesion boundaries across slices.

\textbf{SAM adaptation in medical imaging.}
Prompt-driven modeling, exemplified by the Segment Anything Model (SAM), has become a prominent approach for medical image segmentation. 
Initial adaptations like Med-SA~\cite{wu2025medical} and SAM-Med2D~\cite{cheng2023sam} show that lightweight fine-tuning on medical datasets can achieve competitive multi-organ performance. However, their reliance on manual or heuristic prompts and the lack of integrated semantic reasoning limit their effectiveness in complex scenarios.
To improve semantic guidance, MedCLIP-SAM~\cite{koleilat2024medclip} integrates CLIP-derived embeddings, enabling text-driven segmentation with gScoreCAM~\cite{chen2022gscorecam}. However, the coarse and noisy nature of its language priors compromises localization accuracy. Approaches like TP-DRSeg~\cite{li2024tp} and OMT-SAM~\cite{zhang2025organ} sight better alignment between vision and language embeddings, but they require extensive paired training data and showed limited generalization to unseen variations.
Volumetric extensions such as SAM-Med3D~\cite{wang2023sam} preserve most of SAM's weights while introducing spatial adapters for improved performance on tasks like pancreatic tumor segmentation. Similarly, OpenVocabCT~\cite{li2025towards} leverages radiology reports for open-vocabulary CT segmentation but struggle with aligning text to voxel-level granularity.

The above SAM-based methods all have an inherent defect: It is difficult to fully utilize structured clinical prior knowledge in the entire processing flow.
Our proposed \textbf{\nameofmethod{}} fills this gap by introducing structured, attribute-level priors in a cognitively inspired pipeline, enhancing lesion segmentation performance under inter-class ambiguity and class imbalance.

\section{Method}

To better leverage structured clinical prior knowledge, we propose \nameofmethod{}, a two-stage framework that extracts fine-grained semantic priors (such as organ location, texture, and shape) from clinical text.
This multi-level and fine-grained guidance enhances \nameofmethod{}'s ability to resolve inter-class overlap and improve small-object segmentation accuracy.

\begin{figure}[t]
    \centering
    \includegraphics[width=\textwidth]{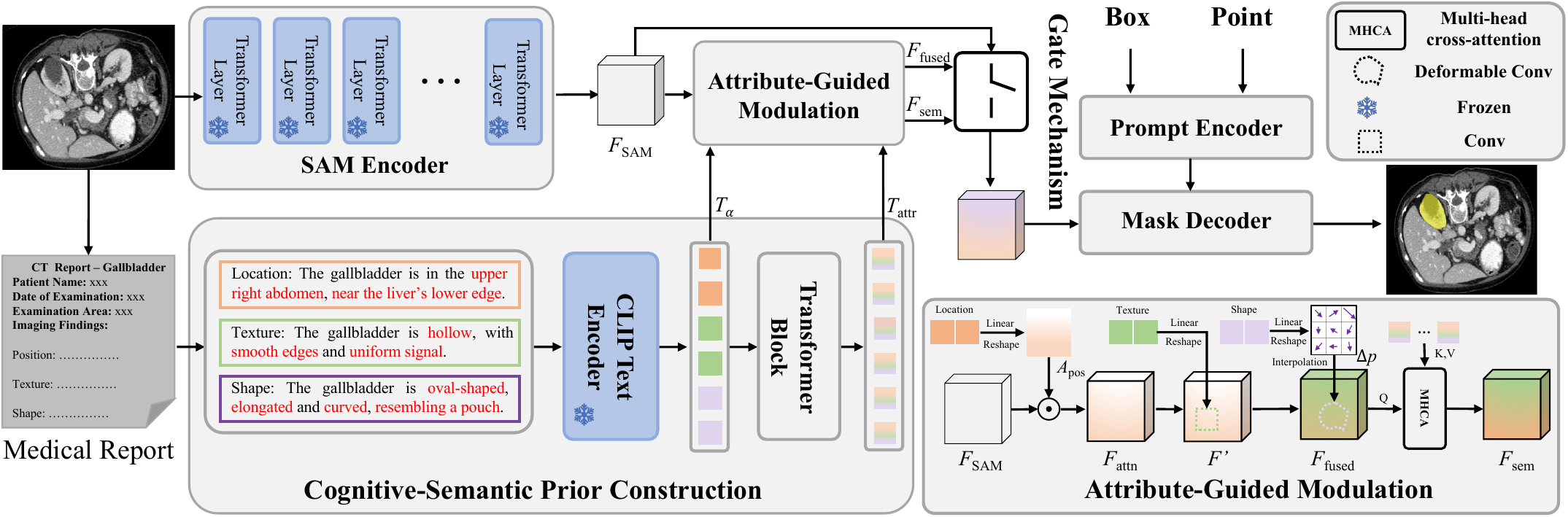}
    \caption{Overview of the~\nameofmethod{} framework.
    The input image features from SAM's encoder are first guided by clinical priors (organ location/texture/shape) in the cognitive stage and then boundary-optimized in the perceptual stage.
    A gating mechanism fuses these refined features with the original SAM features and clinical guidance. The output is fed into SAM's mask decoder with prompt inputs to produce precise segmentation.}
    \label{fig:baseline}
    \vspace{-0.3cm}
\end{figure}

\subsection{Framework of \nameofmethod}

The overall framework of~\nameofmethod{} is illustrated in Fig.~\ref{fig:baseline}.
Given an input image $I$, it is first passed through the SAM image encoder to extract feature representations.
These features are then processed through two key stages of refinement: the \textbf{cognitive} stage and the \textbf{perceptual} stage. 
In the cognitive stage, structured semantic priors, such as organ location, texture, and shape, are extracted and integrated to provide high-level guidance for the feature refinement process.
In the perceptual stage, these priors modulate the feature representations, enabling fine-grained adjustments to the segmentation boundaries.

After passing through these two stages, the refined features, semantic priors, and the output from SAM's image encoder are fused using a gating mechanism.
This mechanism combines the refined features, the SAM encoder output, and the semantic priors to produce an enhanced representation.
Finally, this fused information is combined with the input from the prompt encoder and passed into SAM's mask decoder, which generates the final segmentation result. 
This two-stage refinement process, coupled with the gating mechanism, allows for more accurate segmentation, particularly for small or ambiguous objects, by effectively utilizing structured clinical priors throughout the feature extraction pipeline, which will be shown in the experiment section.

\subsection{Stage I: Cognitive-Semantic Prior Construction.}

In the cognitive stage, we replicate the way clinicians form a high-level understanding of anatomy before closely inspecting the image.
This stage involves extracting key semantic information from clinical reports to guide the subsequent analysis. 
We begin by breaking the clinical report into three key attribute texts that represent critical aspects of anatomy: $T_{\mathrm{pos}}^{\text{txt}}$, $T_{\mathrm{tex}}^{\text{txt}}$, and $T_{\mathrm{shape}}^{\text{txt}}$, which correspond to anatomical location, texture patterns, and object morphology.
These attributes help clinicians understand the lesion’s position, appearance, and form. 
Each attribute text is then encoded using CLIP’s text encoder into token embeddings.
This process mirrors how clinicians mentally categorize key attributes, such as location, texture, and shape, before they begin inspecting the image in detail.

After encoding, the three attribute embeddings are combined and processed using the cross-attribute attention, which can be implemented with a Transformer block to refine each attribute by leveraging the context provided by the others.
For example, a clinician's understanding of the shape of a lesion may influence their interpretation of its texture.
This integrated representation helps to create a more cohesive understanding of the anatomy.
Finally, the fused tokens are aggregated into a single semantic prior vector using average pooling. 
This vector encapsulates the clinical hypothesis, which will guide the subsequent stages of image analysis.
The calculation formulas for this process are as follows:
\begin{align}
T_{\alpha} &= \mathrm{CLIP}_{\mathrm{text}}\bigl(T_{\alpha}^{\text{txt}}\bigr) \;\in\; \mathbb{R}^{L_{\alpha}\times d}, \quad \text{for } \alpha \in \{\mathrm{pos}, \mathrm{tex}, \mathrm{shape}\}, \\T_{\mathrm{cross}} &= \mathrm{TransformerBlock}\bigl([T_{\mathrm{pos}};\,T_{\mathrm{tex}};\,T_{\mathrm{shape}}]\bigr) \;\in\; \mathbb{R}^{(L_{pos}+L_{tex}+L_{shape})\times d}, \\T_{\mathrm{attr}} &= \mathrm{AvePool}\bigl(T_{\mathrm{cross}}\bigr) \;\in\; \mathbb{R}^{1\times d}.
\end{align}
In this formula,  $L_{\alpha}$ represents the sequence length of the input text $T_{\alpha}^{\text{txt}}$, which corresponds to the number of tokens in the text after tokenization, and $d$ refers to the embedding dimension, which is the length of the vector representation for each token generated by the CLIP encoder.
$\mathrm{AvePool}$ represents the average pooling operation.

\subsection{Stage II: Perceptual-Attribute-Guided Modulation}
In the perceptual stage, \nameofmethod{} first uses the features of location, texture, and shape extracted through the CLIP text encoder to guide the output of SAM's image encoder.
Each attribute cue plays a specific role.
Positional cues help focus attention on relevant regions, texture cues adapt the filters to highlight object-specific patterns, and shape cues guide spatial sampling to accurately trace the contours of the target regions.
The guided output is modulated by the semantic prior vector $T_{\mathrm{attr}}$ constructed in the cognitive stage to refine object features.

Given the feature output $F_{\text{SAM}}$ with spatial size $H \times W$ from SAM’s image encoder and $T_{\alpha}, \alpha \in \{\mathrm{pos}, \mathrm{tex}, \mathrm{shape}\} $ from the output of  CLIP text encoder, we first extract the positional prior from $T_{\alpha}$. 
This positional prior $T_{\text{pos}}$ is first flatten and then linearly projected to a spatial attention map $A_{\text{pos}} \in H \times W$ after the reshape operation. Then, $A_{\text{pos}}$ is operated on each channel of $F_{\text{SAM}}$, yielding $F_{\mathrm{attn}}$.
This influences the encoder’s focus, helping suppress irrelevant background responses.
In addition, it ensures that the processing focuses on clinically relevant regions, particularly those that clinicians would typically prioritize.

Next, we use the texture priors $T_{\text{tex}}$ to adjust the resulting $F_{\mathrm{attn}}$, which is tailored to produce imaging characteristics. 
Given $T_{\text{tex}}$, we first flatten it to a 1D vector and then project it to a $7\times7$ matrix via linear projection.
The resulting $7\times7$ matrix is used as a convolutional filter to perform convolution with each channel of $F_{\mathrm{attn}}$, yielding $F'$.
By doing so, \nameofmethod{} amplifies subtle object signals that generic filters might overlook. 
Finally, we use a deformable convolution~\cite{dai2017deformable} to encode the shape priors $T_{\text{shape}}$. $T_{\text{shape}}$ is first flatten and then sent into a linear layer to generate kernels and offsets $\Delta p$, which will then be used in the deformable convolution as done in~\cite{gong2021deformable} to refine spatial sampling according to the expected lesion morphology, yielding $F_{\mathrm{fused}}$.
This step ensures that the receptive fields align with the anticipated object form, which is helpful for improving boundary delineation.
%
%
After guiding the $F_{\mathrm{SAM}}$, we further refine the resulting feature map $F_{\mathrm{fused}}$ using the semantic prior vector $T_{\mathrm{attr}}$ constructed during the cognitive stage.
The refine process can be formulated as follows:
\begin{equation}
    F_{\mathrm{sem}} = \text{MHCA}_{C}(F_{\mathrm{fused}}, F_{\mathrm{attr}}, F_{\mathrm{attr}}),
\end{equation}
where $\text{MHCA}_{C}$ denotes multi-head cross-attention along the channel dimension following~\cite{yin2024camoformer}.




\section{Experiments}


\begin{table}[t]
\centering
\caption{Comparisons between our \nameofmethod{} and other state-of-the-art SAM-based methods on the Med2D-16M dataset. 
We use Dice to evaluate the binary medical segmentation performance.
`*' indicates the experimental results we reproduced.
}
\renewcommand{\arraystretch}{1.0}
\resizebox{1\textwidth}{!}{
\begin{tabular}{c|c|cccccccccc}
\toprule
Prompt  & \textbf{Method} & \textbf{CT} & \textbf{MR} & \textbf{PET} & \textbf{Dermoscopy} & \textbf{Endoscopy} & \textbf{Funds} & \textbf{US} & \textbf{X-ray} &\textbf{Average} \\
\midrule
\multirow{7}{*}{\rotatebox[origin=c]{90}{\centering\textbf{point}}} 
&SAM*~\cite{kirillov2023segment}  & 30.27 & 23.18 & 36.31 & 36.90 & 32.46 & 32.81 & 19.31 & 20.31 &28.97  \\

&SAM2*~\cite{ravi2024sam}  & 63.81 & 44.18 & 71.36 & 83.41 & 58.33 & 74.06 & 64.77 & 46.11 &63.25  \\

&SAM-Med2D*~\cite{cheng2023sam}  &  71.44 & 48.35 & 73.00 & 86.40 & 59.41 & 76.06 &65.00  &49.41  &66.13  \\

&MedSAM*~\cite{ma2024segment} & 73.00 & 51.51 & 68.91 & 87.33 & 55.80 & 77.39 & 69.09 & 53.71 &67.09  \\

&Med-SA*~\cite{wu2025medical} &  73.51 & 46.92 & 73.71 & 89.41 & 60.11 & 77.91 & 66.01 & 42.73 &66.29  \\

&SEG-SAM*~\cite{huang2024seg}  & 78.35 & 53.81 & 77.31 & 88.05 & 69.16 & 80.52 & 71.07 & 60.00 &72.28  \\
&\nameofmethod{} (Ours) &\cellcolor[gray]{.83}80.32 &\cellcolor[gray]{.83}58.91 &\cellcolor[gray]{.83}79.45 &\cellcolor[gray]{.83}90.24 &\cellcolor[gray]{.83}70.74 &\cellcolor[gray]{.83}80.94 &\cellcolor[gray]{.83}72.00 &\cellcolor[gray]{.83}61.53 &\cellcolor[gray]{.83}74.27 \\
\midrule
\multirow{7}{*}{\rotatebox[origin=c]{90}{\centering\textbf{box}}} 

&SAM*~\cite{kirillov2023segment}  & 63.61 & 51.62 & 63.73 & 84.13 & 70.30 & 77.70 & 77.38 & 52.52 &67.62 \\

&SAM2*~\cite{ravi2024sam} & 79.41 & 59.31 & 77.41 & 90.58 & 78.40 & 86.37 & 85.59 & 61.47 &77.32 \\

&SAM-Med2D*~\cite{cheng2023sam} & 81.42 & 61.72 & 79.44 & 92.71 & 77.37 & 87.41 & 85.37 & 59.25 &78.09 \\

&MedSAM*~\cite{ma2024segment} & 83.18 & 63.27 & 77.24 & 90.89 & 74.16 & 84.25 & 85.15 & 64.25 &77.80 \\

&Med-SA*~\cite{wu2025medical} & 83.46 & 61.79 & 80.11 & 92.83 & 77.35 & 85.81 & 84.19 & 57.72 &77.86 \\

&SEG-SAM*~\cite{huang2024seg} & 85.74 & 65.71 & 81.41 & 91.35 & 80.84 & 84.86 & 85.73 &70.16 &80.73\\

&\nameofmethod{} (Ours) &\cellcolor[gray]{.83}86.61 &\cellcolor[gray]{.83}66.94 &\cellcolor[gray]{.83}82.49 & \cellcolor[gray]{.83}92.51 &\cellcolor[gray]{.83}81.73 &\cellcolor[gray]{.83}85.72 &\cellcolor[gray]{.83}87.42 &\cellcolor[gray]{.83}71.74 &\cellcolor[gray]{.83}81.90   \\
\bottomrule
\end{tabular}
}
\label{tab:sam-med2d}
\vspace{-0.3cm}
\end{table}

\subsection{Experimental setup}
\label{experimental}
\textbf{Datasets.} 
We evaluate \nameofmethod{} on SA-Med2D-20M and Synapse.
SA-Med2D-20M~\cite{cheng2023sam} contains multi-modality medical images with dense segmentation masks. 
Following SEG-SAM~\cite{huang2024seg}, we use eight representative modalities for point/box-prompt evaluation.
All modelities are trained on a 25\% subset and tested on the full test set.
To assess generalization without prompts, we also report results on Synapse, a CT dataset with pixel-level annotations for eight abdominal organs.
%

\textbf{Implementation Details.}  
All models are trained for 50 epochs on eight RTX 3090 GPUs using the AdamW optimizer (\(\beta_1{=}0.9\), \(\beta_2{=}0.999\), weight decay \(=0.1\)).  
Input images are uniformly resized to \(224 \times 224\) pixels, with a per-GPU batch size of 12.

\begin{figure}[t]
    \centering
    \includegraphics[width=\textwidth]{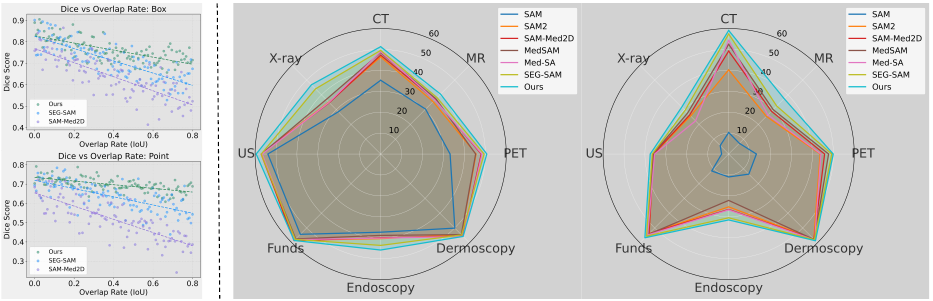}
    \caption{Combined analysis of anatomical overlap robustness (left) and small-object segmentation (right). 
    \textbf{Left:} Dice versus overlap rate under box prompts (top) and point prompts (bottom).
    \nameofmethod{} shows the weakest negative correlation and the flattest regression slope, indicating superior resilience to inter-class ambiguity.
    \textbf{Right:} Radar charts of per-modality Dice scores under box prompts (left) and point prompts (right). 
    \nameofmethod{} consistently outperforms SEG-SAM and other baselines across all eight modalities, with the greatest improvements in challenging, low-contrast domains such as ultrasound and dermoscopy.}
    \label{fig:SAM-Med2D}
    \vspace{-0.3cm}
\end{figure}

\subsection{Comprehensive Evaluation on the SAM-Med2D Dataset}
\label{sam_med2d}

We compare~\nameofmethod{} with seven leading SAM-based segmentation methods on the SAM-Med2D benchmark to assess the benefits of early integration of structured priors, namely location, texture, and shape.
Our evaluation focuses on three aspects: (1) prompt-based performance, (2) robustness to anatomical overlap, and (3) segmentation of very small targets across eight imaging modalities.

\textbf{Prompt-based performance.}  
We first assess performance under  point prompts. Results in the upper half of Table~\ref{tab:sam-med2d} show that \nameofmethod{} achieves an average dice score of \textbf{74.27\%}, outperforming SEG-SAM (72.28\%) by \textbf{+1.99\%}.
Notably, our method leads across all modalities, with the largest improvements observed in MR (+5.10\%) and X-ray (+1.53\%). 
This indicates that even minimal spatial cues can be greatly amplified by structured semantic priors.
Next, we evaluate the models under box prompts, which provide more precise spatial constraints. 
As shown in the lower half of Table~\ref{tab:sam-med2d},~\nameofmethod{} achieves an average Dice score of \textbf{81.90\%}, surpassing SEG-SAM (80.73\%) by \textbf{+1.17\%}.
This improvement further confirms that stronger localization enables more effective exploitation of structural priors. 
Overall,~\nameofmethod{} consistently leverages both sparse and dense prompts to deliver enhanced segmentation accuracy across all eight modalities.

\textbf{Robustness to anatomical overlap.}  
Adjacent anatomical structures often exhibit high visual similarity and spatial proximity, causing inter-class ambiguity and performance degradation.
To quantify robustness, we analyze the correlation between Dice scores and overlap rates for three representative models: \nameofmethod{}, SEG-SAM, and SAM-Med2D, under both box and point prompts.
Results are shown in the left of Fig.~\ref{fig:SAM-Med2D}.
Under the \textbf{box} prompt, \nameofmethod{} exhibits the smallest performance drop, with a Pearson correlation of approximately $r \approx -0.36$ and a regression slope of $-0.12$. 
In comparison, SEG-SAM yields $r \approx -0.58$ (slope $-0.23$), and SAM-Med2D shows $r \approx -0.70$ (slope $-0.36$).
With the \textbf{point} prompt, all methods suffer more due to sparse guidance. However, \nameofmethod{} remains most robust, with $r \approx -0.45$ and slope $-0.15$, compared to SEG-SAM ($r \approx -0.68$, slope $-0.29$) and SAM-Med2D ($r \approx -0.75$, slope $-0.43$).
These consistently weaker negative correlations and shallower slopes confirm the strong ability of \nameofmethod{} to handle overlapping anatomies, validating the effectiveness of early semantic-prior integration.

\textbf{Small object segmentation.}  
Segmenting objects that occupy less than 5\% of the image area is particularly challenging due to severe class imbalance and limited pixel support. 
We systematically evaluate small object performance across eight modalities, using radar plots under both box and point prompts, as shown on the right of Fig.~\ref{fig:SAM-Med2D}.
Under box prompts, \nameofmethod{} achieves an average Dice score of \textbf{51.33\%}, outperforming SEG-SAM (49.57\%) by \textbf{+1.76\%}, with the largest improvement observed in ultrasound (+2.43\%) and dermoscopy (+1.60\%). 
With point prompts, it attains \textbf{59.10\%}, ahead of SEG-SAM (57.41\%) by \textbf{+1.69\%}.
These results highlight the ability of \nameofmethod{} to maintain high sensitivity to small structures, effectively overcoming challenges posed by overlap and class imbalance. 
The radar visualization underscores the consistency and robustness of our approach across diverse imaging modalities.

\begin{table}[t]
\centering
\caption{Comparisons with state-of-the-art models on the Synapse multi-organ CT dataset in both few-shot and fully-supervised settings. 
Greyed values represent our results.
mDice: the Mean Dice coefficient; HD95: the 95th percentile of the Hausdorff Distance.
`*' indicates the experimental results we reproduced.
}
\renewcommand{\arraystretch}{1.1}
\resizebox{1\textwidth}{!}{
\begin{tabular}{c|c|cccccccc|cc}
\toprule
& \textbf{Method} & \textbf{Spleen} & \textbf{Kidney (R)} & \textbf{Kidney (L)} & \textbf{Gallbladder} & \textbf{Liver} & \textbf{Stomach} & \textbf{Aorta} & \textbf{Pancreas} & \textbf{mDice} $\uparrow$ & \textbf{HD95} $\downarrow$ \\
\midrule

\multirow{5}{*}{\rotatebox[origin=c]{90}{\centering\textbf{Non-SAM}}} 
& TransUnet~\cite{chen2021transunet}    & 87.23 & 63.13 & 81.87 & 77.02 & 94.08 & 55.86 & 85.08 & 75.62 & 77.48 & 31.69 \\
& SwinUnet~\cite{cao2022swin}            & 85.47 & 66.53 & 83.28 & 79.61 & 94.29 & 56.58 & 90.66 & 76.60 & 79.13 & 21.55 \\
& TransDeepLab~\cite{azad2022transdeeplab} & 86.04 & 69.16 & 84.08 & 79.88 & 93.53 & 61.19 & 89.00 & 78.40 & 80.16 & 21.25 \\
& DAE-Former~\cite{azad2023dae}          & 88.96 & 72.30 & 86.08 & 80.88 & 94.98 & 65.12 & 91.94 & 79.19 & 82.43 & 17.46 \\
& MERIT~\cite{rahman2024multi}           & 92.01 & 84.85 & 87.79 & 74.40 & 95.26 & 85.38 & 87.71 & 71.81 & 84.90 & 13.22 \\
\midrule
\multirow{5}{*}{\rotatebox[origin=c]{90}{\centering\textbf{SAM}}} 
& AutoSAM~\cite{hu2023efficiently}    & 80.54 & 80.02 & 79.66 & 41.37 & 89.24 & 61.14 & 82.56 & 44.22 & 62.08 & 27.56 \\
& SAM Adapter~\cite{chen2023sam}       & 83.68 & 79.00 & 79.02 & 57.49 & 92.68 & 69.48 & 77.93 & 43.07 & 72.80 & 33.08 \\
& SAMed~\cite{zhang2023customized}       & 87.33 & 80.10 & 82.75 & 70.24 & 93.37 & 73.62 & 86.99 & 67.64 & 80.26 & 28.89 \\
& H-SAM~\cite{cheng2024unleashing}       & 92.34 & 85.99 & 87.71 & 69.65 & 95.20 & 86.27 & 87.53 & 72.53 & 84.65 & 7.29 \\
& PG-SAM~\cite{zhong2025pg} & 93.12 & 84.57 & 87.93 & 73.26 & 95.40 & 86.62 & 87.87 & 71.49 & 84.79 & 7.61 \\
&SEG-SAM* &92.10 &84.16 &86.39 &70.21 &94.91 &86.06 &87.42 &71.48 &84.09 &10.26\\
& \nameofmethod{} (Ours)   &\cellcolor[gray]{.83}93.49 &\cellcolor[gray]{.83}86.70 &\cellcolor[gray]{.83}89.71 &\cellcolor[gray]{.83}78.29 &\cellcolor[gray]{.83}95.17 &\cellcolor[gray]{.83}87.77 &\cellcolor[gray]{.83}88.91 &\cellcolor[gray]{.83}73.71 &\cellcolor[gray]{.83}86.71 &\cellcolor[gray]{.83}6.13  \\
\bottomrule
\end{tabular}
}
\label{tab:synapse}
\vspace{-0.3cm}
\end{table}

\begin{figure}[t]
    \centering
    \begin{overpic}[width=\textwidth]{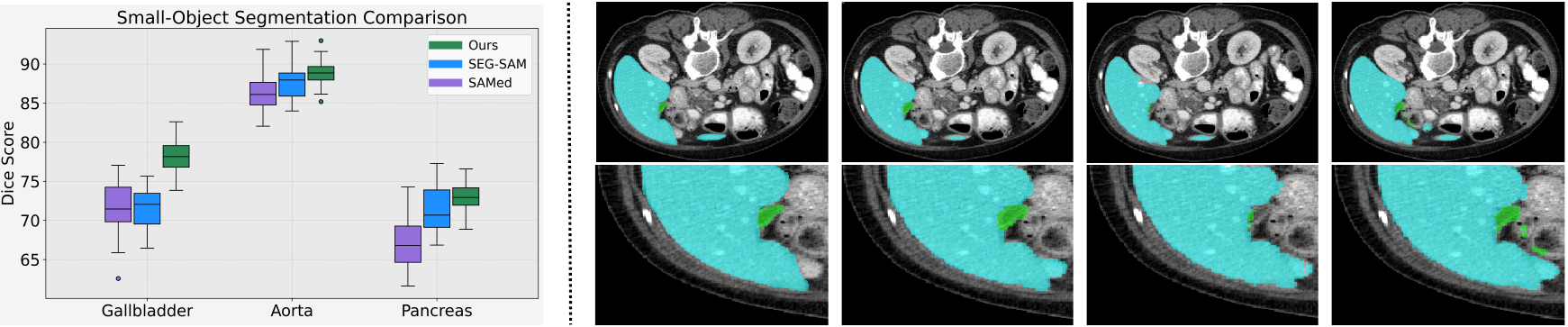}
    \put(44, -3){{\textcolor{black}{GT}}}
    \put(60, -3){{\textcolor{black}{Our}}}
    \put(71, -3){{\textcolor{black}{SEG-SAM}}}
    \put(89, -3){{\textcolor{black}{SAMed}}}
  \end{overpic}
  \vspace{0.05cm}
    \caption{\textbf{Left}: Box plot of dice scores for three small volume organs, gallbladder, pancreas, and aorta, on the Synapse dataset.
    Each distribution compares \nameofmethod{}, SEG-SAM, and SAMed under identical settings, with \nameofmethod{} showing consistently higher medians and tighter spreads, highlighting its superior sensitivity to small-object segmentation amid class imbalance and anatomical overlap. 
   \textbf{Right}: Segmentation results on the Synapse dataset, showing masks of the gallbladder (green) and liver(blue) generated by different methods, illustrating the improved boundary delineation of \nameofmethod{}. illustrating \nameofmethod{}’s improved boundary delineation.
    }
    \label{fig:synapse}
    \vspace{-0.3cm}
\end{figure}
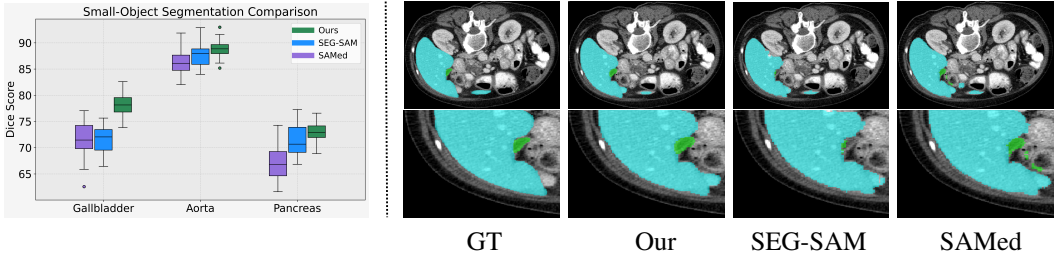

\subsection{Evaluation on Synapse: Multi-Organ CT Segmentation.}
\label{synapse}

To assess the performance of~\nameofmethod{} in fully automatic segmentation settings, we do experiments on the Synapse dataset, which contains contrast-enhanced abdominal CT scans annotated for eight organ types.
Unlike previous evaluations relying on point or box prompts, this prompt-free setup enables direct evaluation of the model’s intrinsic segmentation capability without external guidance.

\textbf{Overall segmentation performance.}  
We compare~\nameofmethod{} with several state-of-the-art fully supervised and SAM-based methods, including SEG-SAM, H-SAM, and PG-SAM.
As shown in Table~\ref{tab:synapse},~\nameofmethod{} achieves the highest mean Dice score of \textbf{86.71\%} and the lowest HD95 of \textbf{6.13}, outperforming SEG-SAM (84.09\%) by \textbf{+2.62\%} and PG-SAM (84.79\%) by \textbf{+1.92\%}.
Notably, our method yields consistent improvements for all organs, with substantial gains on anatomically challenging structures, such as the \textbf{gallbladder} (+5.03\%) and \textbf{pancreas} (+2.22\%).
These results demonstrate the sustained effectiveness of early-integrated structured semantic priors, even in scenarios without user guidance.

\textbf{Robustness to anatomical overlap.}  
To evaluate the performance in spatially entangled regions, we examine organ pairs known for strong anatomical overlap and boundary ambiguity, including \textit{gallbladder-liver}, \textit{pancreas-stomach}, and \textit{kidney-aorta}.
These structures are prone to inter-class confusion due to partial occlusion and weak boundary contrast.
As shown in Table~\ref{tab:synapse},~\nameofmethod{} consistently outperforms competing methods on challenging structures, achieving Dice scores of 78.29\% for the gallbladder, 73.71\% for the pancreas, and 87.77\% for the stomach.
This indicates that our semantic guidance mechanism enhances spatial discrimination and improves robustness in complex anatomical environments.

\textbf{Small object segmentation.}  
Although the Synapse dataset primarily contains medium-to-large abdominal organs, structures such as the \textbf{gallbladder}, \textbf{pancreas}, and \textbf{aorta} remain particularly challenging due to their small spatial extent, high class imbalance, and frequent boundary overlap with neighboring tissues.
We designate these organs as representative small structures and illustrate their segmentation performance using box plots shown on the left of Fig.~\ref{fig:synapse}.

\nameofmethod{} achieves both the highest median Dice scores and the most compact performance distributions for all three organs, outperforming SEG-SAM by +8.08\% on the gallbladder, +2.23\% on the pancreas, and +1.49\% on the aorta.
These results demonstrate the model’s enhanced sensitivity to small anatomical targets and its robustness under limited pixel supervision and inter-class interference.
The right of Fig.~\ref{fig:synapse} further illustrates the benefits of structured semantic guidance. 
The figure  compares segmentation results across methods for the gallbladder. 
\nameofmethod{} achieves clearer boundary delineation and more accurate mask alignment, underscoring the effectiveness of early semantic prior integration in small object segmentation.

\begin{table}[t]
  \footnotesize
  \begin{minipage}{0.47 \textwidth}
  \centering
  \setlength{\tabcolsep}{2.8mm}
\renewcommand\arraystretch{1.1}
\caption{
    Ablation study of structured semantics at the cognitive stage.
    We evaluate the individual and combined contributions of location, texture, and shape.
    "Loc." and "Tex." denote location and texture.
    Results are reported as Dice scores on the KiTS19 and Kvasir datasets. 
    Best results are highlighted in gray.
    }
  \begin{tabular}{lcc|cccccc}
    \toprule

    Loc. &Tex.  &Shape  &KiTS19 &Kvasir  \\
    \midrule
       & & &77.21 &76.46\\
       \cmark & & &83.91 &83.39 \\
       &\cmark & &82.99 &82.96  \\
      & &\cmark &82.25 &82.07  \\
      \midrule
    \cmark&\cmark & &86.38 &85.71    \\
    \cmark& &\cmark &85.14 &84.34    \\
    &\cmark &\cmark &85.81 &84.69 \\
    \midrule
     \cmark &\cmark &\cmark &\cellcolor[gray]{.83}88.23 &\cellcolor[gray]{.83}87.69 \\
    \bottomrule
  \end{tabular}
  \label{tab:attribute}
  \end{minipage}
  \hfill
\begin{minipage}{0.5\textwidth}
  \centering
  \renewcommand\arraystretch{1.1}
  \setlength{\tabcolsep}{1.2mm}
    \caption{
    Ablation study of structured semantic priors at the perceptual stage.
    We evaluate the individual and combined contributions of location, texture, and shape priors.
    "Loc." and  "Tex." denote location and texture.
    Results are reported as Dice scores on the KiTS19 and Kvasir datasets. 
    Best results are highlighted in gray.
    }
  \begin{tabular}{lcc|cccccc}
    \toprule

    Loc. prior &Tex.prior  &Shape prior  &KiTS19 &Kvasir  \\
    \midrule
    & & &80.39 &84.01\\
    \cmark & & &85.97 &85.31 \\
    &\cmark & &85.09 &84.74  \\
      & &\cmark &83.79 &84.11  \\
      \midrule
    \cmark&\cmark & &87.77 &86.99    \\
    \cmark& &\cmark &87.39 &86.59    \\
    &\cmark &\cmark &86.01 &84.89 \\
    \midrule
     \cmark &\cmark &\cmark &\cellcolor[gray]{.83}88.23 &\cellcolor[gray]{.83}87.69 \\
    \bottomrule
  \end{tabular}
\label{tab:perceptual}
\end{minipage}
\vspace{-0.4cm}
\end{table}

\subsection{Ablation Study}

We perform extensive ablation experiments to validate the effectiveness of the proposed modules, including structured semantic priors, the two-stage architecture, cross-attribute attention, and the gating mechanism. All results are reported as Dice scores on the KiTS19 and Kvasir datasets.

\textbf{Effect of structured semantics at the cognitive stage.}
Table~\ref{tab:attribute} analyzes the impact of structured semantics (location, texture, and shape) when incorporated at the cognitive stage. Each type of semantic prior individually improves performance over the baseline (77.21 / 76.46). Among them, location priors contribute the most (83.91 / 83.39), while texture and shape bring moderate gains. Combining two types of priors leads to further improvements, and the best performance (88.23 / 87.69) is achieved when all three priors are jointly used. 

\textbf{Effect of structured semantics at the perceptual stage.}
As shown in Table~\ref{tab:perceptual}, introducing structured priors into the perceptual stage also consistently improves performance. Location priors again play a dominant role (85.97 / 85.31). Combining two priors shows stronger results, and the full combination of all priors leads to the best performance (88.23 / 87.69). This indicates that structured semantics can also effectively guide low-level visual representation learning.

\textbf{Effectiveness of the two-stage framework and cross-attribute attention.}
We further evaluate the contribution of the cognitive stage, the perceptual stage, and the cross-attribute attention (CAA) in Table~\ref{tab:two_stage}. Removing both stages leads to the lowest performance (77.21 / 76.46). Enabling either stage improves results, with the perceptual stage providing stronger gains than the cognitive stage. Combining both stages achieves the highest score (88.23 / 87.69), confirming their complementarity. Notably, the introduction of CAA brings a clear improvement (from 87.22 / 86.14 to 88.23 / 87.69), demonstrating its ability to facilitate interaction between diverse structured priors.

\textbf{Analysis on the gating mechanism.}
Finally, Table~\ref{tab:gate_attention} investigates the contribution of each feature branch and their combinations in the fusion stage. The fused feature $F_{\mathrm{fused}}$, cognitively refines the feature $F_{\mathrm{sem}}$, and the SAM feature $F_{\mathrm{sam}}$, each individually improving performance over the baseline. Combining any two of them further enhances accuracy, and the full gating of all three features yields the best performance (88.23 / 87.69). This confirms the effectiveness of the gating mechanism in selectively integrating complementary information from multiple feature streams.

\begin{table}[t]
  \footnotesize
  \begin{minipage}{0.4\textwidth}
  \centering
  \setlength{\tabcolsep}{2.1mm}
  \renewcommand\arraystretch{1}
  \caption{
    Ablation on the effectiveness of the cognitive stage (Cog.), perceptual stage (Per.), and cross-attribute attention (CAA) in~\nameofmethod{}. 
    We examine their individual and combined contributions on KiTS19 and Kvasir datasets. 
    `CAA' denotes the cross-attribute attention module used within the cognitive stage.
    Results are reported as dice scores on the KiTS19 and Kvasir datasets.
    Best results are highlighted in gray.
  }
 \begin{tabular}{lc|c|cccccc}
    \toprule
    Cog. &Per. &CAA &KiTS19 &Kvasir  \\
    \midrule
    & & &77.21 &76.46 \\
    \cmark   & & &80.39 &84.01 \\
        &\cmark & &86.05 &85.20 \\
     \cmark  & \cmark & &\cellcolor[gray]{.83}88.23 &\cellcolor[gray]{.83}87.69 \\
     \midrule
     \cmark & \cmark & &87.22 &86.14 \\
     \cmark &\cmark &\cmark &\cellcolor[gray]{.83}88.23 &\cellcolor[gray]{.83}87.69 \\
    \bottomrule
  \end{tabular}
  \label{tab:two_stage}
  \end{minipage}
  \hfill
\begin{minipage}{0.55\textwidth}
  \centering
  \setlength{\tabcolsep}{3.5mm}
  \renewcommand\arraystretch{1.2}
    \caption{
    Ablation study on the gating mechanism for feature fusion. 
    We evaluate the individual and combined effects of three feature branches: the perceptually guided feature $F_{\mathrm{fused}}$, the cognitively refined feature $F_{\mathrm{sem}}$, and the raw feature from SAM $F_{\mathrm{sam}}$.
    The results are reported in terms of dice score on the KiTS19 and Kvasir datasets.
    Best performance is highlighted in gray.
    }
  \begin{tabular}{lcc|cccccc}
    \toprule
  $F_{fused}$ &$F_{sem}$& $F_{sam}$&KiTS19 &Kvasir  \\
    \midrule
    \cmark  & & &85.03 &84.15 \\
        &\cmark & &85.91 &84.67 \\
      & & \cmark &85.60 &84.40 \\
        \midrule
     \cmark & \cmark & &86.71 & 85.69\\
     \cmark  & &\cmark   &86.11 &85.31 \\
      &\cmark  &\cmark &87.30 &86.72  \\
      \midrule
     \cmark &\cmark &\cmark &\cellcolor[gray]{.83} 88.23 &\cellcolor[gray]{.83}87.69 \\
    \bottomrule
  \end{tabular}
\label{tab:gate_attention}
\end{minipage}
\vspace{-0.3cm}
\end{table}

\section{Conclusions}
\label{conclusions}
In this work, we present \nameofmethod, a structured semantic attribute-guided framework that enhances SAM’s segmentation capabilities in the presence of small lesions and inter-class anatomical overlap. 
By introducing position-, texture-, and shape-aware guidance into the feature encoding and fusion process, \nameofmethod{} enables more precise delineation of subtle lesion boundaries and improves robustness in structurally ambiguous regions.
Comprehensive experiments across multi-organ and lesion-centric benchmarks demonstrate that our method outperforms existing SAM-based and universal segmentation baselines, particularly in low-contrast and imbalanced settings.

\textbf{Limitations.}  
While our method integrates structured priors to improve segmentation robustness, these priors are still learned implicitly from data rather than being explicitly encoded.  
As a result, the model may struggle to fully adapt to patient-specific anatomical variations or rare pathological patterns.  
This challenge is common across existing SAM-based frameworks. However,~\nameofmethod{} demonstrates stronger resilience in such scenarios by incorporating a clinically inspired reasoning mechanism, leading to improved generalization compared to prior approaches.
Future work could incorporate explicit clinical knowledge or case-specific adaptation to further enhance robustness to rare or out-of-distribution cases.



{\small
 \bibliographystyle{plainnat} 
\bibliography{neurips_2025}
}

\appendix

\section{Appendix}
\subsection{Overview}
In this appendix, we provide additional results and details to supplement the main paper.
The content is organized as follows:

\begin{itemize}
   
    \item \textbf{Dataset and Implementation Details:} In Sec.~\ref{experiment_setup}, we include further information on dataset partitioning and pre-processing protocols, as well as key training settings such as optimizer parameters.

    \item \textbf{Structured Semantic Attribute Extraction:} In Sec.~\ref{semantic_attributes}, we describe the process for extracting structured semantic attributes from medical reports generated by LLM.

    \item \textbf{Computational Efficiency Comparison:} In Sec.~\ref{computational_efficiency}, We present quantitative results on model efficiency in terms of trainable parameter counts and inference speed (FPS), demonstrating the advantage of our design.

    \item \textbf{Ablation Studies:} In Sec.~\ref{ablation_study}, additional experiments are conducted to evaluate the impact of dataset size, model capacity, and the number of point prompts on segmentation performance.
    
    \item \textbf{Visualization Results:} In Sec.~\ref{visualization_results}, additional segmentation results across various modalities are presented to further demonstrate the effectiveness and generalizability of \nameofmethod{}.
\end{itemize}

\begin{figure}[t]
    \centering
    \includegraphics[width=\textwidth]{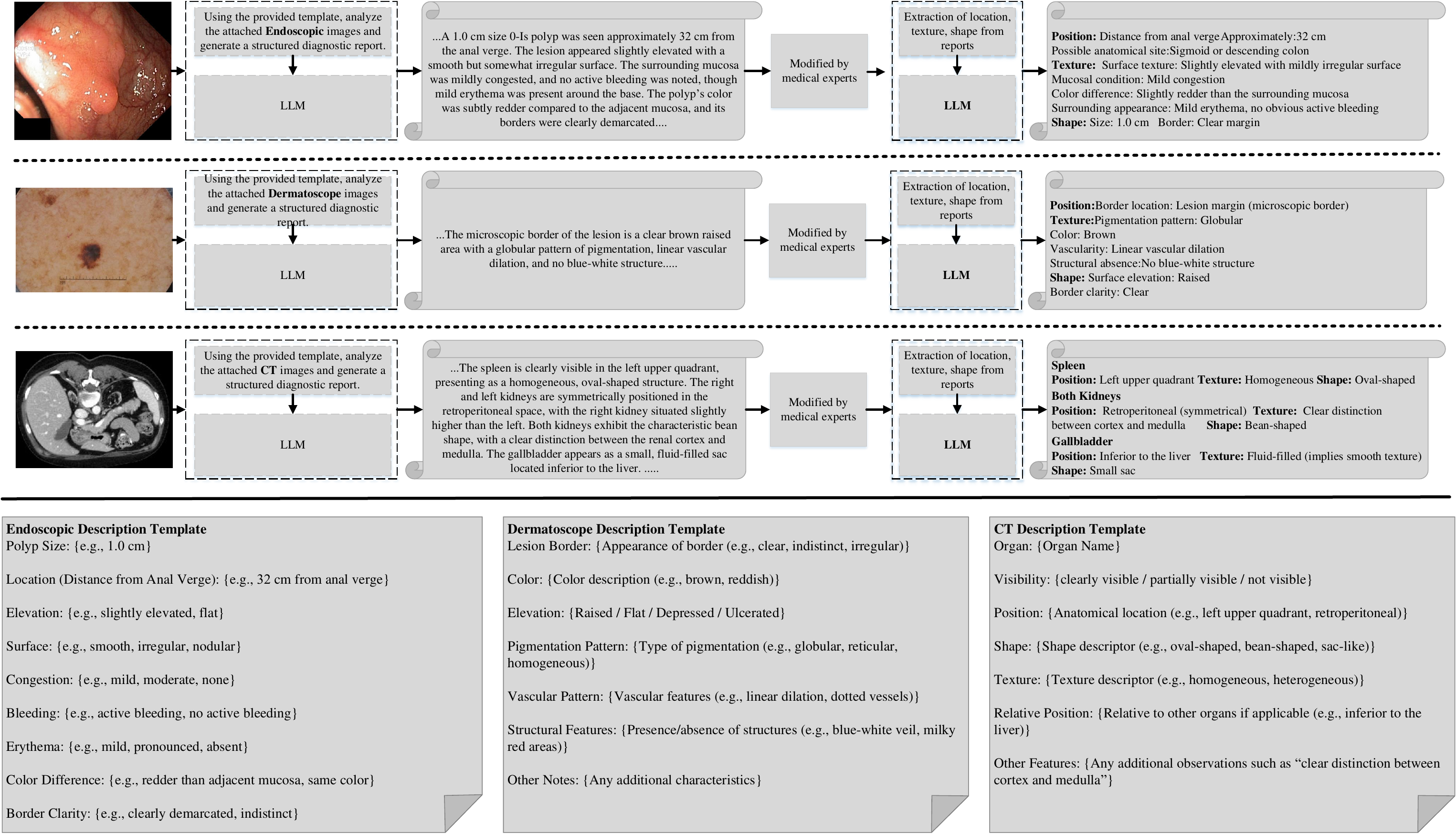}
    \caption{An illustration of the process for extracting structured semantic attributes from medical images across three modalities: CT, endoscopy, and dermoscopy.
    The \textbf{upper} part of the figure shows the pipeline from image input to language-based attribute extraction. 
    The \textbf{lower} part of the figure demonstrates how prompts are constructed based on the input image, which are then used by the LLM to generate the medical report.}
    \label{fig:report}
\end{figure}

\begin{figure}[t]
    \centering
    \includegraphics[width=\textwidth]{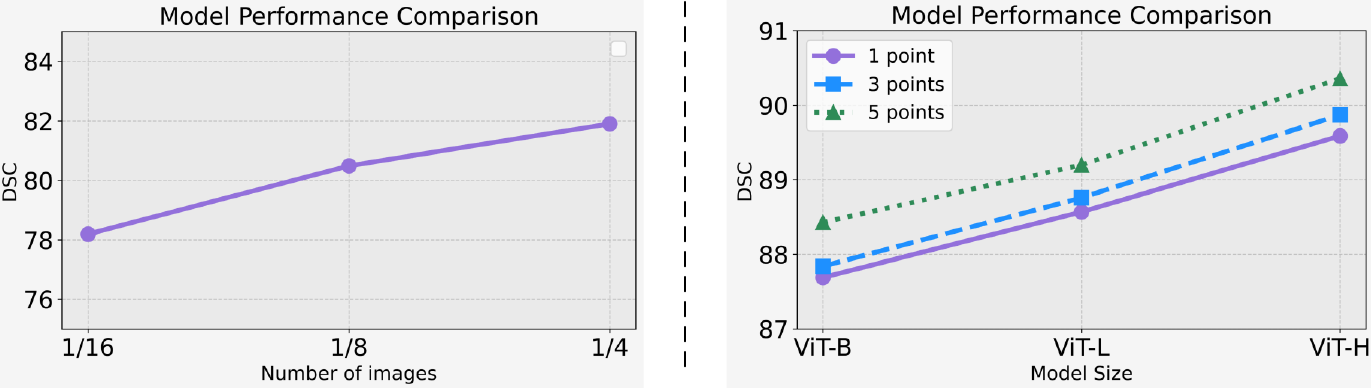}
    \caption{Impact of data volume and prompt strategy on performance.
    \textbf{Left:} Comparison of model performance under different training data volumes(e.g., 1/16, 1/8, and 1/4).
    \textbf{Right:} Evaluation across model scales (ViT-B, ViT-H, ViT-L) with varying point-based prompts (e.g., 1-point, 2-points, and 3-points). 
    Our method exhibits consistent gains with increased model capacity.}
    \label{fig:medical_size}
\end{figure}

\subsection{Experiment setup}
\label{experiment_setup}
\textbf{Datasets.} We report the results on SA-Med2D-20M and Synapse.
SA-Med2D-20M introduced by SAM-Med2D~\cite{cheng2023sam} contains 4.6 million medical images paired with approximately 20 million segmentation masks spanning ten imaging modalities.
In line with SEG-SAM~\cite{huang2024seg}, we select eight clinically representative modalities for our point- and box-prompt experiments: CT, MR, PET, US, X-ray, endoscopy, dermoscopy, and fundus.
All data are divided into 80\% for training and 20\% for testing.
Due to computational constraints, both the \nameofmethod{} and baseline models are trained on a randomly sampled 25\% subset of the training data, while the evaluation is performed on the entire test dataset.
To evaluate the generalizability of~\nameofmethod{} under prompt-free conditions, we further conduct experiments on the Synapse dataset.
This dataset comprises 3,779 contrast-enhanced abdominal CT slices with pixel-wise annotations for eight major organs, offering a consistent single-modality setting.
Unlike SA-Med2D-20M, no prompts are provided during training or inference, enabling us to rigorously assess segmentation performance—particularly in anatomically complex and small object scenarios.

\textbf{Implementation Details.}  
All models are trained for 50 epochs on eight RTX 3090 GPUs using the AdamW optimizer (\(\beta_1{=}0.9\), \(\beta_2{=}0.999\), weight decay \(=0.1\)).  
Input images are uniformly resized to \(224 \times 224\) pixels, with a per-GPU batch size of 12.
We follow the data pre-processing strategy of SAM-Med2D~\cite{cheng2023sam}.
For 3D images, all slices along the three orthogonal axes are extracted and treated as independent 2D samples. 
For 2D images, we normalize pixel values to the range [0, 255]
To reduce label noise, we discard ground truth masks occupying less than 0.153\% of the image area, as they are too small to extract meaningful features.

\begin{table}[t]
\centering
\caption{Comparison of tuned parameters and inference speed (FPS) among SAM-based segmentation methods.
\nameofmethod{} (Ours) achieves significantly higher FPS than most baselines while reducing the number of trainable parameters, demonstrating the efficiency of our structured semantic guidance.
}
\renewcommand{\arraystretch}{1.0}
\resizebox{0.7\textwidth}{!}{
\begin{tabular}{cccccccccccc}
\toprule
 \textbf{Method} &Model  & \textbf{Tuned Params(M)} &FPS\\
\midrule
SAM~\cite{kirillov2023segment} &ViT-B    &89.88 &8   \\

SAM2~\cite{ravi2024sam} &hiera-base-plus   &80.8 &64.8  \\

SAM-Med2D~\cite{cheng2023sam}  &ViT-B  &163 &35    \\

MedSAM~\cite{ma2024segment}   &ViT-B   &93.73   &13    \\

\nameofmethod{} (Ours)  &ViT-B    &68.49    &48 \\
\bottomrule
\end{tabular}
}
\label{tab:compute}
\end{table}

\subsection{Extracting Structured Semantic Attributes}
\label{semantic_attributes}
The pipeline for extracting structured semantic attributes is illustrated in Fig.~\ref{fig:report}.
Given an input image (e.g., CT, endoscopy, or dermoscopy), a large language model (LLM) is first employed to generate a preliminary descriptive medical report.
This report is subsequently reviewed and refined by medical experts to ensure clinical accuracy and consistency.
The revised report is then re-processed by the LLM to extract fine-grained semantic attributes, specifically the position, texture, and shape of relevant anatomical or pathological structures.
These structured attributes serve as lightweight yet informative priors that can be seamlessly integrated into downstream segmentation or diagnostic models, enhancing both interpretability and generalizability across imaging modalities.

\subsection{Computational efficiency}
\label{computational_efficiency}
To evaluate computational efficiency, we compare our method \nameofmethod{} with representative SAM-based models in terms of the number of tuned parameters and inference speed (FPS), as shown in Table~\ref{tab:compute}. 
While the original SAM~\cite{kirillov2023segment} and MedSAM~\cite{ma2024segment} fine-tune a large number of parameters (over 89M), our method freezes the SAM backbone and only tunes structured semantic guidance and mask decoder, resulting in a significantly smaller trainable parameter count (68.49M).
In addition, \nameofmethod{} achieves an inference speed of 48 FPS, outperforming MedSAM (13 FPS) and SAM-Med2D (35 FPS). 
These results highlight the efficiency of our approach in balancing performance and computational cost, making it more suitable for real-time or resource-constrained clinical applications.

\subsection{Ablation study}
\label{ablation_study}

\textbf{Impact of Dataset Size on Model Performance.}
To assess the effect of training data volume on model performance, we conduct experiments using different proportions of the SAM-Med2D dataset, specifically 1/16, 1/8, and 1/4 of the training dataset.
As shown in the left of Fig.~\ref{fig:medical_size}, model performance—as measured by Dice Similarity Coefficient (DSC)—increases steadily with the number of training images. 
Even with only 1/16 of the data, our model achieves a DSC of 78.19\%, and this performance rises to 81.90\% when trained on 1/4 of the dataset. 
These results highlight the scalability of our approach and its ability to effectively learn from increasing amounts of training data, without any changes to supervision strategies.

\textbf{Effect of Model Size and Prompt Quantity on Segmentation Performance.}
We evaluate the performance of models with different capacities (ViT-B, ViT-L, ViT-H) under varying numbers of point prompts (1, 3, and 5) using the SAM-Med2D dataset. 
As shown in the right of Fig.~\ref{fig:medical_size}, while larger models consistently yield better segmentation results—as measured by Dice Similarity Coefficient (DSC)—the performance improvement from increasing the number of prompts is relatively modest across all model sizes.
For instance, ViT-H achieves a DSC of 89.59\% with a single point, and only improves to 90.36\% with five points.
This trend suggests that our model is capable of generating accurate segmentation with minimal user input, demonstrating strong robustness even in low-prompt or prompt-free scenarios.
This property makes the model particularly suitable for real-world clinical settings where extensive annotation is often impractical.

\begin{figure}[t]
    \centering
    \begin{overpic}[width=1\textwidth]{{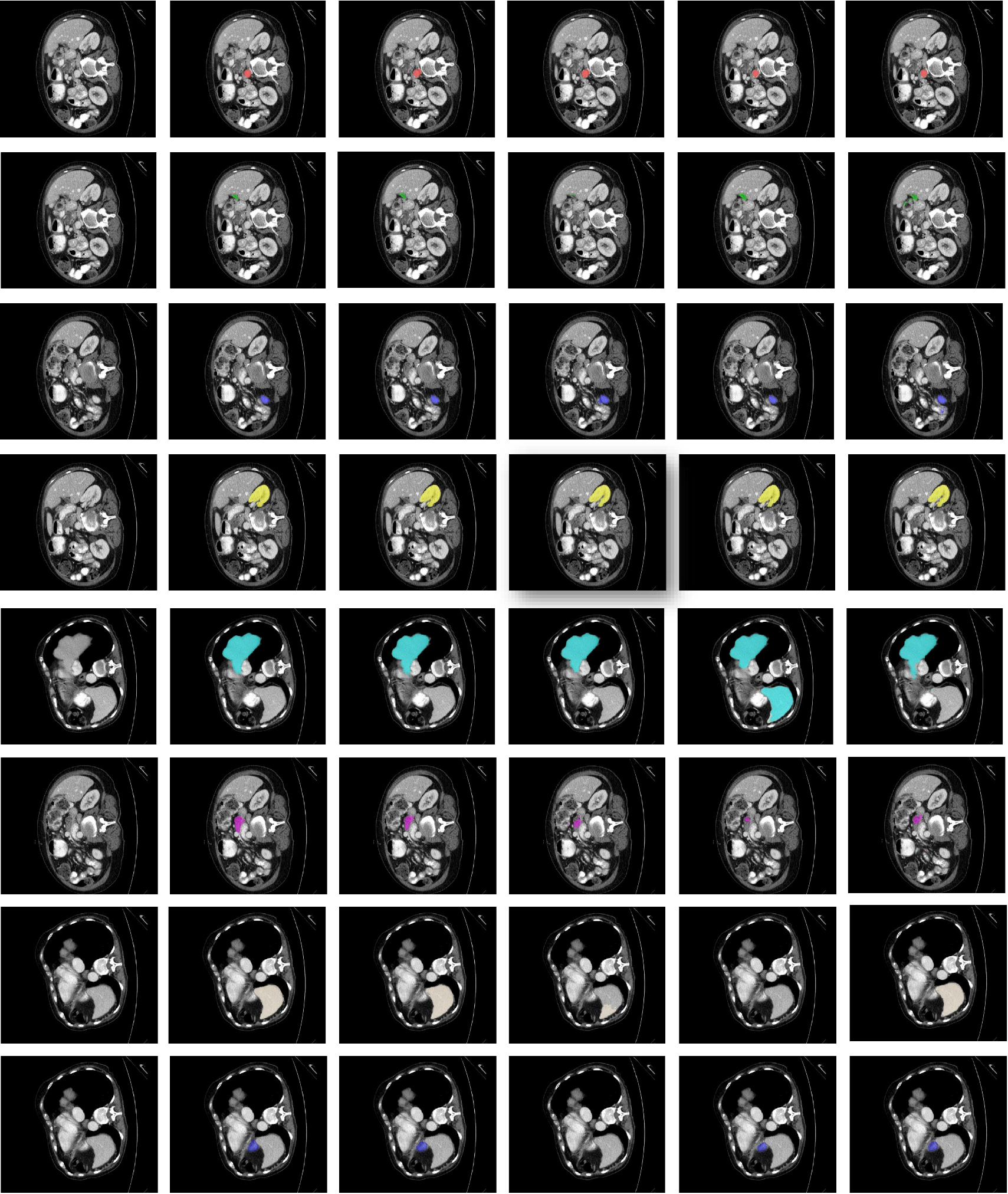}}
    \put(4, 101){{\textcolor{black}{image}}}
    \put(20, 101){{\textcolor{black}{GT}}}
    \put(33, 101){{\textcolor{black}{Our}}}
    \put(46, 101){{\textcolor{black}{SAM}}}
    \put(59, 101){{\textcolor{black}{MedSAM}}}
    \put(73, 101){{\textcolor{black}{SAM-Med2D}}}

    \put(-2,92){\rotatebox{90}{Aotra}}
    \put(-2,77){\rotatebox{90}{Gallbladder}}
    \put(-2,64){\rotatebox{90}{Kidney(R)}}
    \put(-2,52){\rotatebox{90}{Kidney(L)}}
    \put(-2,40){\rotatebox{90}{Liver}}
    \put(-2,28){\rotatebox{90}{Spleen}}
    \put(-2,14.5){\rotatebox{90}{Pancreas}}
    \put(-2,2){\rotatebox{90}{Stomach}}

  \end{overpic}
    \caption{Segmentation results of different methods on the Synapse dataset.}
    \label{fig:synapse_visual}
\end{figure}

\begin{figure}[t]
    \centering
    \begin{overpic}[width=1\textwidth]{{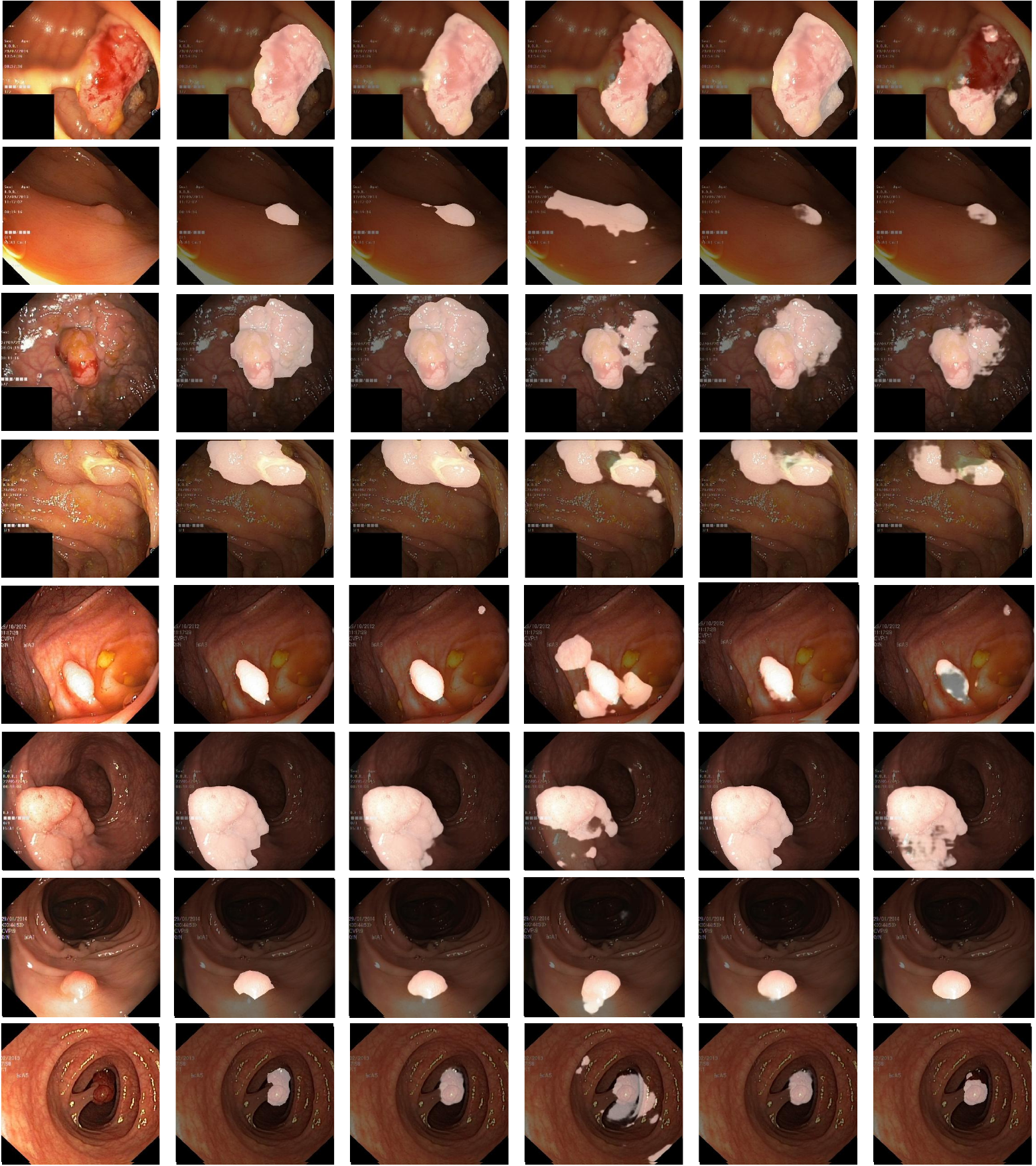}}
    \put(4, 101){{\textcolor{black}{image}}}
    \put(21, 101){{\textcolor{black}{GT}}}
    \put(34.5, 101){{\textcolor{black}{Our}}}
    \put(49.5, 101){{\textcolor{black}{SAM}}}
    \put(60.5, 101){{\textcolor{black}{SAM-Med2D}}}
    \put(77.5, 101){{\textcolor{black}{MedSAM}}}
  \end{overpic}
    \caption{Segmentation results of different methods on the Endoscopy modality.}
    \label{fig:endoscopic_visual}
    \vspace{-0.3cm}
\end{figure}

\begin{figure}[t]
    \centering
    \begin{overpic}[width=1\textwidth]{{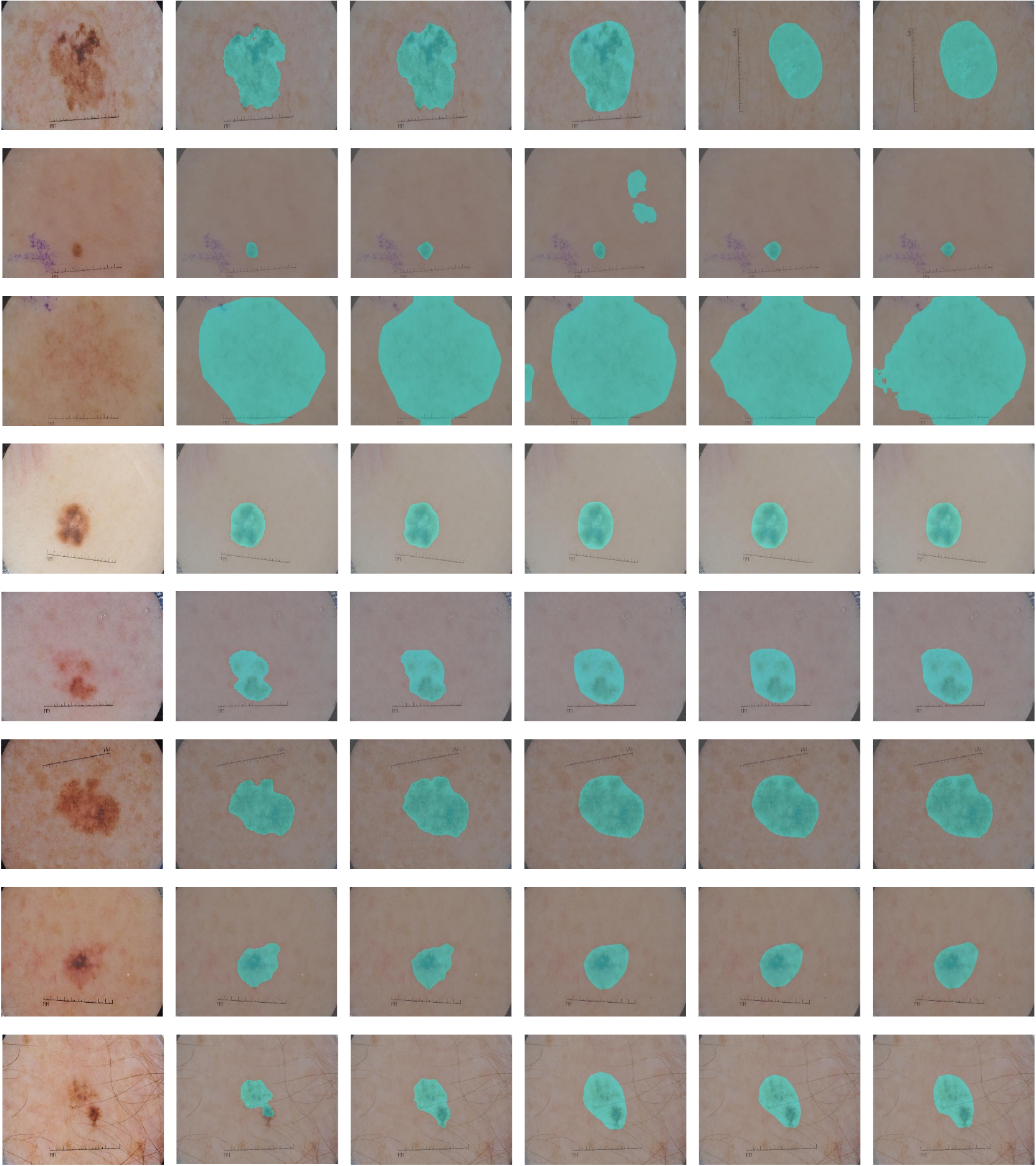}}
    \put(4, 101){{\textcolor{black}{image}}}
    \put(21, 101){{\textcolor{black}{GT}}}
    \put(34.5, 101){{\textcolor{black}{Our}}}
    \put(49.5, 101){{\textcolor{black}{SAM}}}
    \put(60.5, 101){{\textcolor{black}{SAM-Med2D}}}
    \put(77.5, 101){{\textcolor{black}{MedSAM}}}
  \end{overpic}
    \caption{Segmentation results of different methods on the Derscopy modality.}
    \label{fig:dermoscop_visual}
    \vspace{-0.3cm}
\end{figure}

\subsection{Visualization results}
\label{visualization_results}

To qualitatively assess the segmentation capability of~\nameofmethod{} across diverse imaging modalities, we present representative results on three different types of medical images: computed tomography (CT), dermoscopy, and endoscopy.
These modalities differ significantly in terms of anatomical structures, visual appearance, and imaging conditions, providing a rigorous evaluation of the generalizability of our approach.

As shown in Fig.~\ref{fig:synapse_visual},  Fig.~\ref{fig:endoscopic_visual} , and  Fig.~\ref{fig:dermoscop_visual}, our method produces accurate and consistent segmentation masks across all three modalities. 
For CT images,~\nameofmethod{} is able to delineate organ boundaries with high precision, even in cases with low contrast or ambiguous edges.
In dermoscopic images, which often exhibit high intra-class variation and low inter-class contrast, our method captures the fine-grained lesion structures while preserving clear margins. 
Similarly, in endoscopic scenarios, where occlusions, lighting variations, and texture complexity pose significant challenges, \nameofmethod{} demonstrates robust performance in segmenting irregular and often small anatomical targets.

These results highlight the strong adaptability of our method to both multi-modality ( Fig.~\ref{fig:endoscopic_visual} and  Fig.~\ref{fig:dermoscop_visual}) and prompt-free (Fig.~\ref{fig:synapse_visual}) settings.
Importantly, the structured semantic priors incorporated into \nameofmethod{} contribute to more anatomically plausible predictions, especially in visually complex scenes where traditional prompt-driven models often struggle. 
We also provide visual comparisons with representative baselines, including SAM~\cite{kirillov2023segment}, SAM-Med2D~\cite{cheng2023sam}, and MedSAM~\cite{ma2024segment}, to further substantiate the effectiveness of \nameofmethod{}.



\end{document}